\documentclass[runningheads]{llncs}
\usepackage{amsmath}
\usepackage{amssymb}
\usepackage{booktabs}
\usepackage[T1]{fontenc}
\usepackage{graphicx}
\usepackage{multirow}
\usepackage{makecell}
\usepackage{subfig}
\usepackage{tipa}

\begin{document}

\title{When Smaller Wins: Dual-Stage Distillation and Pareto-Guided Compression of Liquid Neural Networks for Edge Battery Prognostics}

\titlerunning{\textsc{DLNet}: Dual-Stage Distillation Liquid Neural Networks}

\author{Dhivya Dharshini Kannan\inst{1} \and Wei Li\inst{1} \and
Wei Zhang\inst{1}\thanks{Corresponding author.} \and \\Jianbiao Wang\inst{2} \and Zhi Wei Seh\inst{2} \and Man-Fai Ng\inst{3}}

\authorrunning{D. D. Kannan et al.}

\institute{Singapore Institute of Technology (SIT), Singapore 828608\\
\email{\{dhivyadharshini.kannan, wei.li, wei.zhang\}@singaporetech.edu.sg}
\and
Institute of Materials Research and Engineering (IMRE), Agency for Science, Technology and Research (A*STAR), 2 Fusionopolis Way, Innovis \#08-03,\\Singapore 138634, Republic of Singapore
\email{\{wang\_jianbiao, sehzw\}@a-star.edu.sg}
\and
Institute of High Performance Computing (IHPC), Agency for Science, Technology and Research (A*STAR), 1 Fusionopolis Way, \#16–16 Connexis,\\Singapore 138632, Republic of Singapore
\email{ngmf@a-star.edu.sg}
}

\maketitle            

\begin{abstract}
Battery management systems increasingly require accurate battery health prognostics under strict on-device constraints. This paper presents \textsc{DLNet}, a practical framework with dual-stage \underline{d}istillation of \underline{l}iquid neural \underline{net}works that turns a high-capacity model into compact and edge-deployable models for battery health prediction. \textsc{DLNet} first applies Euler discretization to reformulate liquid dynamics for embedded compatibility. It then performs dual-stage knowledge distillation to transfer the teacher model’s temporal behavior and recover it after further compression. Pareto-guided selection under joint error–cost objectives retains student models that balance accuracy and efficiency. We evaluate \textsc{DLNet} on a widely used dataset and validate real-device feasibility on an Arduino Nano 33 BLE Sense using \texttt{int8} deployment. The final deployed student achieves a low error of 0.0066 when predicting battery health over the next 100 cycles, which is 15.4\% lower than the teacher model. It reduces the model size from 616 kB to 94 kB with 84.7\% reduction and takes 21 ms per inference on the device. These results support a practical \textit{smaller wins} observation that a small model can match or exceed a large teacher for edge-based prognostics with proper supervision and selection. Beyond batteries, the \textsc{DLNet} framework can extend to other industrial analytics tasks with strict hardware constraints. Code is available at:
\url{https://github.com/Dhivya-DD17/DLNet}

\end{abstract}

\section{Introduction}
\label{sec:intro}
Electrified transport is scaling rapidly. Global electric vehicle (EV) sales reached about 14 million in 2023 and grew further to around 17 million in 2024 \cite{IEA25}. As EV adoption scales, battery performance and health increasingly determine operation safety, reliability, and the total cost of modern EV systems as battery degradation affects range, failure risk, and maintenance cost. In practice, related systems and functionalities rely on battery management systems (BMSs) to monitor and protect the battery pack, where accurate health forecasting supports longevity and operational safety under real driving conditions. Within battery prognostics and health management, state-of-health (SoH) and remaining useful life (RUL) are widely treated as key indicators, which help reduce risk and cost by modelling and predicting degradation before battery's end-of-life (EoL). Meanwhile, real BMS deployment remains highly constrained by embedded compute and memory budgets, so practical solutions must be not only accurate for modelling and prediction, but also compact and exportable to edge devices. 

Machine learning (ML) has been one of the dominant approaches for battery health prognostics. ML models learn degradation patterns from historical data and use them to predict future battery dynamics. Representative advancements include time-series models such as long short-term memory (LSTM) \cite{kim2021forecasting} and liquid neural networks (LNNs) \cite{li2026entrolnn}, attention and transformer-based models \cite{zhu2024sparse}, and recent solutions using generative artificial intelligence \cite{kwan5871665knowdiff}. These models demonstrate competitive prediction accuracy under offline evaluation. However, edge deployability, e.g., model size and runtime compatibility, is often not treated as a design constraint. As a result, many high-performing models remain difficult to use in resource-limited battery management systems, despite their academic merits. To improve deployability, model compression is a typical pathway with several techniques such as knowledge distillation, pruning, and quantization. Beyond single-technique compression, existing studies also adopt hybrid solutions based on multiple techniques, or update the optimization target with practical constraints. Recently, battery-specific compression has attracted attention for lightweight prognostics. For example, distillation is used to compress the computational-intensive denoising into a lightweight model for battery health prediction \cite{ye2025simple}, and create small models that require less sensors for battery measurements \cite{xie2024knowledge}. While these studies provide valuable evidence that compression reduces model complexity, they provide limited discussion of end-to-end embedded executability, and consequently, there remains a practical gap between model compression and resource-constrained deployment for battery prognostics.

In this paper, we design and develop a deployment-aware framework for battery health prognostics and propose \textsc{DLNet} \textipa{/di:l nEt/}, dual-stage \underline{d}istillation and Pareto-guided compression of \underline{l}iquid neural \underline{net}works. \textsc{DLNet} aims to compress a high-performing LNN teacher model into compact students with competitive prediction performance through a carefully designed dual-stage pipeline. We first train a high-capacity LNN teacher model in an accuracy-first manner and treat the trained model as a knowledge source. To enable embedded deployment, we then create a diverse pool of students by reformulating the teacher’s liquid dynamics into Euler-based discretized variants with different model configurations. The students are trained via distillation with guidance from both the teacher and ground-truth labels, and we select elite students using Pareto analysis under two objectives for prediction error and deployment cost. To meet strict device constraints, \textsc{DLNet} further compresses the selected elites via pruning with different sparsity levels and introduces another round of distillation, followed by Pareto selection to retain the most practical students. Finally, \textsc{DLNet} applies quantization to the selected students, exports them to an embedded inference format, and completes deployment.
We validate \textsc{DLNet} on a real embedded device, Arduino Nano 33 BLE Sense, one of the most compact Arduino boards that supports on-device ML inference. Notably, the final deployed student is much smaller while achieving high prediction accuracy. \textsc{DLNet} achieves a mean absolute error (MAE) of 0.0066 for forecasting SoH over 100 future cycles, which is a 15.4\% error reduction compared to the teacher. It also reduces model size from 616 to 94 kB, i.e., an 84.7\% size reduction. This real-device deployment supports a practical \textit{smaller wins} observation in \textsc{DLNet}, where the teacher provides a stable target during training, compression alleviates noise sensitivity, and Pareto-based selection helps pick optimal models from a diverse student pool. Overall, \textsc{DLNet} offers an effective deployment-oriented framework for turning accurate battery ML models into edge deployments with optimized error–cost trade-offs. Beyond battery health prognostics, the framework can be customized and extended to other battery tasks such as battery capacity estimation and also to non-battery applications that face similar embedded constraints.

The remainder of this paper is organized as follows. We first we present the workflow and technical details of \textsc{DLNet} in Section \ref{sec:method}. Section \ref{sec:experiment} provides experimental results and discussions. Finally, Section \ref{sec:conclusion} concludes the paper.

\section{DLNet Methodology}
\label{sec:method}
\textsc{DLNet} is a deployment-aware framework for edge battery prognostics. Fig. \ref{fig:sys} provides an overview of \textsc{DLNet}, and the technical details are presented below.

\begin{figure}[t]
\centering
\includegraphics[width=0.925\textwidth]{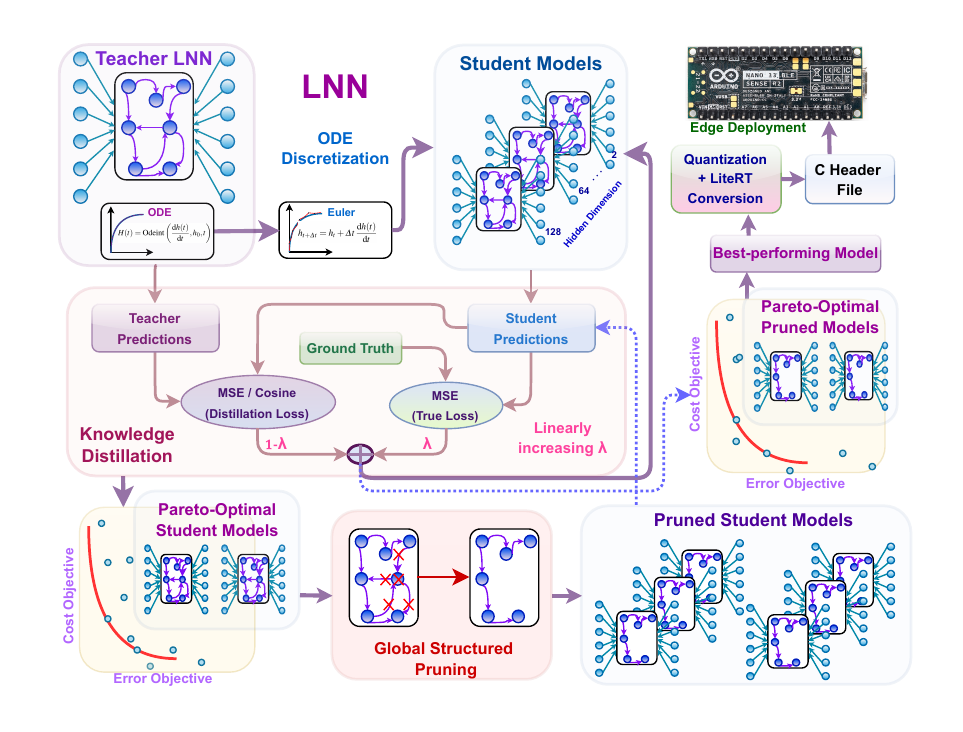}
\caption{An illustration of the system architecture of \textsc{DLNet} with two compression, distillation, and model selection stages followed by edge deployment.}
\label{fig:sys}
\end{figure}

\subsection{LNN Teacher Model}
The teacher model in \textsc{DLNet} is a high-capacity LNN. The choice is motivated by existing works (e.g., \cite{li2026entrolnn}), which have demonstrated the effectiveness of liquid dynamics in modelling long-horizon and non-stationary battery health patterns. A detailed empirical justification is provided later in the experiments, where the LNN teacher is benchmarked against alternative ML teachers. In \textsc{DLNet}, the LNN teacher is trained in an accuracy-first manner and then frozen entirely to serve as a knowledge source for student learning. Its continuous-time state dynamics are modeled through a learned ordinary differential equation (ODE) system that encodes diverse battery aging behaviors into latent representations, which are decoded into multi-cycle SoH forecasts and later transferred to compact students. Formally, let $\tau$ denote the number of observed historical cycles used as the input window. At the current cycle $c$, the input SoH sequence is defined as $X = [x_{c-\tau+1}, \ldots, x_c]$. The teacher model $\mathcal{T}_\theta(\cdot)$ is configured as a sequence-to-sequence temporal forecaster to forecast the SoH over the horizon of next $\tau'$ cycles as $\hat{Y} = [\hat{y}_{c+1}, \ldots, \hat{y}_{c+\tau'}] = \mathcal{T}_\theta(X)$, where $\theta$ represents the teacher parameters and the teacher is trained to minimize a standard regression loss between $\hat{Y}$ and the ground-truth before student distillation. The input window length $\tau$ and forecast horizon length $\tau'$ are independent design choices in LNN, and we configure $\tau=\tau'$ in this study for narrative simplicity. 

The teacher model $\mathcal{T}_\theta(\cdot)$ contains three conceptual blocks. The first block is the encoder, a temporal embedding module that maps raw battery cycle measurements into a latent representation of degradation behavior. The encoder first expands the input through a trainable linear projection, followed by \texttt{LayerNorm}, and applies \texttt{ReLU} activation and dropout to stabilize gradients. The normalized latent is then compressed again through a second linear projection, followed by another \texttt{LayerNorm} and a \texttt{Tanh} activation, producing the hidden state, which encodes the smooth and drifting patterns typical in battery health signals. Second, we have the dynamics block. To provide temporal knowledge, the teacher evolves the hidden state in continuous time, governed by a learned ODE as,
\begin{equation}
    \frac{\mathrm{d} h(t)}{\mathrm{d} t} = -\alpha h(t) + \beta \tanh\big(W h(t) + \bar{u}\big),
\end{equation}
where $h(t)$ is the hidden state, $\alpha$ and $\beta$ are trainable parameters that control the decay and nonlinearity, respectively, $W$ is the learnable dynamics matrix, and $\bar{u}$ is the mean of the input $X$. This ODE is integrated once per inference to obtain the evolved latent state $h_T$ as,
\begin{equation}
    h_T = \texttt{odeint} \Big(\frac{\mathrm{d} h(t)}{\mathrm{d} t}, h_0, t \in [0, T]\Big),
\end{equation}
where $T$ denotes the endpoint of integration. In practice, the state dynamics block is realized as a deep chain of liquid recurrent neurons that iteratively refine the fixed-dimensional latent state. Each neuron contains a trainable linear operator learned during teacher training, and the ODE integrator solves the stacked recursion in one pass to produce the enriched latent state $h_T$, which is later decoded for battery health forecasts. The evolved state is next processed by the decoder as the last block of the teacher model. First, the state is expanded into a fixed-dimension representation, followed by \texttt{LayerNorm}, \texttt{ReLU} activation, and a \texttt{dropout} layer. Then, we further compress the processed representation with another linear layer, followed by \texttt{LayerNorm} and \texttt{ReLU} activation. Finally, the last linear layer maps the latent representation into $\tau'$ predicted future SoH within the forecast horizon. Overall, the teacher is a high-capacity model that serves as a reliable knowledge source for student learning. While the teacher model design itself is not a novelty we emphasize, the learned predictive behavior is critical and preserved to enable downstream distillation, detailed as follows.

\subsection{First-Stage Distillation and Pareto Selection}
The teacher model, while accurate, relies on continuous-time ODE solving that is heavy for edge deployment with constrained computing resources and repeated inference tasks. To make the solution practical, we aim to create lightweight models that are cheaper to run at edge. However, simply compressing a teacher model often reduces model performance significantly. Here, \textsc{DLNet} generates many students first, which later are improved through distillation.

\paragraph{Euler-based Students Creation}
Discretization is a natural way to simplify inference and improve deployment efficiency. \textsc{DLNet} approximates the LNN teacher’s continuous-time hidden-state dynamics with an explicit Euler discretization, which replaces iterative ODE-solver-based integration at inference with a solver-free and learnable state-update. This reformulation preserves the teacher’s liquid-transition behavior while producing lightweight student models. Specifically, each student mimics the teacher’s state evolution via,
\begin{equation}
    h(t+\Delta t) = h(t) + \Delta t \cdot \frac{\mathrm{d} h(t)}{\mathrm{d} t},
\end{equation}
where $h(t)$ is the hidden state. Substituting the teacher's ODE dynamics into the Euler step gives the explicit student update as,
\begin{equation}
    h(t+\Delta t) = h(t) + \Delta t \cdot \Big(-\alpha h(t) + \beta \tanh\big(W h(t) + \bar{u}\big)\Big),
\end{equation}
where $W$ is the teacher trained dynamics matrix that students imitate, $\alpha$ and $\beta$ are inherited parameters, and $\bar{u}$. To enhance the embedded inference, the student dynamics matrix $W'$ is parameterized in a low-rank and diagonal-stable form as,
\begin{equation}
    W' = \operatorname{diag}(w_{\text{diag}}) + \frac{1}{r} U V^\intercal, 
\end{equation}
where $w_{\text{diag}}$ is a learnable vector and $U$ and $V$ form a rank-$r$ residual term with low-rank factor size $r<d$ for hidden state dimension $d$. As a result, the parameter count is reduced from from $O(d^2)$ to $O(d + dr)$, e.g., $O(d)$ given a constant $r$, and the student inference can be substantially cheaper. 

We would like to mention that Euler discretization is not only a design choice for efficiency and it resolves a real engineering deployment constraint. A widely-used embedded inference standard is LiteRT which enables portable and optimized model execution on low-power deices. LiteRT, however, does not support general ODE solvers or iterative numerical integration operators, so a continuous-time teacher model cannot be exported directly to LiteRT for edge deployment. Explicit Euler-based discretization reformulates the hidden-state evolution into a single solver-free tensor update that preserves the learnable knowledge while remaining compatible with LiteRT export. To build a diverse pool of students, \textsc{DLNet} generates many Euler-based students that keep the same forecasting interface as the teacher while varying the hidden dimension $d$. We choose different dimensions as powers of 2, e.g., 64 and 128, so as to align with memory and vectorized compute units on embedded processors. Each student is independently initialized with dynamic parameters and the large pool of students increases the likelihood of finding compact models that perform strongly. 

\paragraph{Student Training with Distillation}
In the created students, Euler is only an approximation of the teacher' ODE dynamics and there can be a non-trivial performance gap. So, we apply knowledge distillation to transfer the teacher's complex liquid dynamics adequately into the Euler-based students with properly trained parameters for each student. Here, each student is trained using a joint loss that balances ground-truth regression and teacher guidance with,
\begin{equation}
\label{eq:loss}
    \mathcal{L}_{\text{total}} = \lambda\, \mathcal{L}_{\text{true}} + (1 - \lambda)\, \mathcal{L}_{\text{distill}},
\end{equation}
where $\mathcal{L}_{\text{total}}$ is the total training loss which includes the true loss $\mathcal{L}_{\text{true}}$ based on the ground-truth labels and distillation loss $\mathcal{L}_{\text{distill}}$ compared to the teacher model. The importance of the two loss terms is controlled by a parameter $0<\lambda<1$. In this study, we update $\lambda$ in each epoch to reflect a practical learning priority. At the beginning, we emphasize teacher-guided learning, where the teacher model provides stable and smooth knowledge signals to help students capture general degradation trends early. Ground-truth data often contains irregular noises with negative impacts such as unstable gradients and overfitting, so it is less focused at the beginning. Eventually, students shift trust from teacher guidance to true labels in place. After learning reasonable liquid transition patterns, students start to specialize based on the ground-truth data. In this study, we configure the upper and lower bounds of $\lambda$, which is initialized in the first epoch and updated in each epoch as $\lambda := \lambda + \Delta \lambda$,
where $\Delta \lambda$ is the increase amount per epoch. Furthermore, the distillation loss term $\mathcal{L}_{\text{distill}}$ can be configured using either \texttt{MSE} or \texttt{Cosine} loss. For each student, we train one student variant per loss type, which doubles the size of the student pool to improve model diversity and distilled behavior coverage. 

\paragraph{Pareto-Guided Student Selection}
After distillation, we select one or more trained students from the student pool, in a way that reflects deployment-aware learning goals rather than accuracy alone. Accuracy measures prediction correctness but not practicality, e.g., models with similar accuracy can have very different resource demands. To preserve both correctness and practicality, we evaluate distilled students in a bi-objective error-cost trade-off space defined by,
\begin{equation}
\label{eq:bi-obj}
    f_{\texttt{err}}(\mathcal{S}) = \sum\nolimits_{i=1}^{n^{\texttt{err}}} w_i^{\texttt{err}} \varepsilon_i(\mathcal{S}),\quad f_{\texttt{cst}}(\mathcal{S}) = \sum\nolimits_{j=1}^{n^{\texttt{cst}}} w_j^{\texttt{cst}} c_j(\mathcal{S}),
\end{equation}
where $f_{\texttt{err}}(\cdot)$ and $f_{\texttt{cst}}(\cdot)$ are the objective functions for prediction errors and model costs, respectively, and the input of both functions is a model $\mathcal{S}$. Battery health prognostics is a complex task, and the model performance cannot be simplified to one aspect only. Here, we design each objective function to cover complementary aspects and gain the flexibility to balance priorities of each aspect through weighting. We assume $n^{\texttt{err}}$ aspects for the error function. The error of aspect $i$ is represented by $\varepsilon_i(\mathcal{S})$ for a model $\mathcal{S}$ and the weight of this error is $w_i^{\texttt{err}}$. Similarly, the cost function covers $n^{\texttt{cst}}$ aspects, where $c_j(\mathcal{S})$ is the cost of cost aspect $j$ for model $\mathcal{S}$ and the corresponding weight is $w_j^{\texttt{cst}}$. 

The error function covers $n^{\texttt{err}}=5$ aspects. First, we consider MAE, root mean square error (RMSE), and mean absolute percentage error (MAPE) to measure average forecast correctness across the cycles in the forecast horizon. Furthermore, the error aspects include uncertainty and confidence to reflect forecast reliability. The former is computed as the mean standard deviation (SD) over multiple, e.g., 100, inference runs, and the latter measures empirical coverage of the 95\% prediction interval across the multiple runs. The cost objective includes $n^{\texttt{cst}}=4$ aspects, including model size, inference time, energy efficiency, and CO2 emission. The first two quantify model deployment feasibility in edge devices, and the rest two capture sustainability impact which is a big concern of current research communities. The values of all aspects are normalized independently so they have comparable impact on the objective functions. Overall, incorporating weighted and minimized terms improves student diversity without relying on a single metric which may not be realistic. 

We identify the most promising elite students for next stage distillation and eventually edge deployment. First, let $\mathcal{P}$ be the pool of trained students. We first introduce two practical thresholds $f_{\texttt{err}}^{\max}$ and $f_{\texttt{cst}}^{\max}$, and exclude students that are either too inaccurate or too expensive for edge deployments. We have,
\begin{equation}
    \mathcal{P} \leftarrow \mathcal{P}\setminus \{\mathcal{S}\mid f_{\texttt{err}}(\mathcal{S})> f_{\texttt{err}}^{\max} \lor f_{\texttt{cst}}(\mathcal{S})> f_{\texttt{cst}}^{\max}\}.
\end{equation}
With the updated student pool, we adopt Pareto analysis to select students with balanced error and cost. This Pareto-based selection is inspired by genetic algorithms, where elite individuals are kept as the strongest population subset for the following evolutionary search. Here, a student $\mathcal{S}$ is defined as Pareto-optimal if no other student $\mathcal{S}'$ dominates it in both objectives, i.e.,
\begin{equation}
\label{eq:pareto}
    \mathcal{S} ~(\text{dominates}) \prec \mathcal{S}'~  \iff~ [f_{\texttt{err}}(\mathcal{S}) < f_{\texttt{err}}(\mathcal{S}')] \land [f_{\texttt{cst}}(\mathcal{S}) < f_{\texttt{cst}}(\mathcal{S}')].
\end{equation}
Then, we update the pool by removing all students who are dominated by at least one student. The resulting one or multiple Pareto-optimal students form the updated pool and serve as the knowledge carriers distilled from the teacher.

\subsection{Second-Stage Distillation}

\paragraph{Model Pruning}
The Euler-discretized students are not compact enough for practical edge deployment and we apply parameter pruning to further reduce model footprint. In this work, pruning follows a standard magnitude-based strategy, where low-importance parameters or neuron connections are removed based on ranking criteria derived from the trained weights. This step directly addresses the constraint of achieving compact models while maintaining LiteRT compatibility. We generate multiple variants per student with different pruning sparsity levels for practical student pool diversity. The resulting pruned students preserve the same forecasting input–output interface and latent degradation patterns.

\paragraph{Distillation and Pareto Selection}
After pruning, student models may have shifted prediction behaviors with reduced parameters. So, we introduce another round of distillation to strengthen students by recovering and stabilizing their useful knowledge learned from the teacher. Here, we reuse the same procedure applied in the first distillation. Each student is trained to minimized the loss in Eq. (\ref{eq:loss}) with the same update rule of the weight $\lambda$. Both \texttt{MSE} and \texttt{Cosine} losses are considered for the distillation loss term, so the student pool size is doubled after distillation. We evaluate all retrained students using the two objectives for prediction error and cost. The elite student subset is updated by keeping only the Pareto-optimal student models based on Eq. (\ref{eq:pareto}) and these models are expected to offer both high prediction accuracy and deployment efficiency.

\paragraph{Edge Deployment}
To deploy the selected student model on compact edge devices, we convert each model into a LiteRT graph using \texttt{int8} quantization during export, remove unsupported operators, and generate a C file containing model weights and structure for embedded inference. The model input is pre-scaled and cast to \texttt{int8} before inference, and the output is de-quantized and rescaled back to the original SoH cycle range. This export-time quantization and solver-free graph conversion ensure LiteRT execution and edge deployment feasibility.

\section{Experiments}
\label{sec:experiment}

\subsection{Experiment Setup}
The well-known MIT-Stanford battery cycle dataset \cite{severson2019data} is used in this study for model training and evaluation with real lithium-ion battery SoH degradation patterns. Training sequences are constructed from historical cycles, and the test set consists of SoH profiles spanning high, medium, and low health ranges, to enable model performance assessment across different scales of battery lifespan. The training and test data are split in the ratio 80:20.

For each teacher or student model, the input window length is 100 cycles and the forecast horizon length is also 100 cycles, i.e., $\tau=\tau’=100$. Each model is trained for 200 epochs. The control parameter $\lambda$ in the loss function starts at 0.1 and increases by 0.004 per epoch until it reaches 0.9. We evaluate performance in terms of prediction errors and deployment costs. Errors are measured by MAE, RMSE, and MAPE computed over the predicted SoH values within the forecast horizon. Prediction uncertainty is quantified from 100 repeated inferences as the mean of the per-run SD, and confidence is measured as one minus the empirical coverage of the 95\% prediction interval. For the error utility $f_{\texttt{err}}$, we assign 50\% total weight to MAE, RMSE, and MAPE (equally shared), and 50\% total weight to uncertainty and confidence (equally shared). For the cost utility $f_{\texttt{cst}}$, we assign 50\% weight to model size, 20\% to inference time, and the remaining 30\% equally shared by energy and CO2 emission, measured using EmissionsTracker from CodeCarbon. Two thresholds used for selecting practical Pareto-optimal students are $f_{\texttt{err}}^{\max} = 0.25$ and $f_{\texttt{cst}}^{\max} = 0.25$. All models are trained on a workstation with an NVIDIA RTX 4050 GPU, and dataset-based validation is conducted inside a Docker environment to ensure reproducibility of results across hardware setups.

\subsection{Teacher Model Foundation}
In the domain of battery health prognostics, many ML models have been studied and demonstrate competitive performance in servers, but fewer works verify if compressed models remain executable and well-performed on edge devices with tight constraints such as model size. In this part, we examine different ML models' compatibility and performance for edge-based battery prognostics. We evaluate several commonly used ML algorithms for battery analytics including convolutional neural networks (CNNs), temporal convolutional networks (TCNs), LSTM, gated recurrent unit (GRU), Transformer, and the LNN with the same hyper-parameters, e.g., 2 hidden layers and a hidden dimension of 128. We consider three implementations for each ML algorithm, including PyTorch for the teacher, and two quantized versions exported with LiteRT in \texttt{float16} and \texttt{int8}. This experiment focuses on whether a model can be reliably exported and executed on edge devices and qualifies as a teacher for \textsc{DLNet}. For each implementation, we evaluate predictive quality and Table \ref{tab:teacher} shows the results.

\begin{table*}[t]
\caption{Comparison of ML models implemented in PyTorch, LiteRT with \texttt{float16}, and LiteRT with \texttt{int8} under two minimized objectives. Error metrics include MAE, RMSE, and MAPE (\%). Cost metrics include inference time (ms) and model size (kB). Model eligibility for error requires MAE $\leq 0.01$. Model eligibility for cost requires \texttt{int8} format and a quantized model size $\leq 500$ kB.}
\centering
\renewcommand{\arraystretch}{1.1}
\setlength{\tabcolsep}{5pt}
\resizebox{0.98\textwidth}{!}{%
\begin{tabular}{c|c|rrr|rr|cc}
\hline\hline
\multirow{1}{*}{Model} & \multirow{1}{*}{Version} & \multicolumn{1}{c}{MAE} & \multicolumn{1}{c}{RMSE} &
\multicolumn{1}{c|}{MAPE} &
\multicolumn{1}{c}{Inference} &
\multicolumn{1}{c|}{Size} &
\multirow{1}{*}{Error} &
\multirow{1}{*}{Cost} \\\hline\hline

\multirow{3}{*}{CNN} & PyTorch & 0.03997 & 0.04148 & 4.59535 & 1.569 & 1418.1 & $\times$ & $\times$ \\ \cline{2-9}
& LiteRT \texttt{float16} & 0.00659 & 0.00721 & 0.76039 & 0.541 & 373.9 & \checkmark & $\times$ \\ \cline{2-9}
& LiteRT \texttt{int8} & 0.01298 & 0.01337 & 1.42713 & 0.471 & 371.3 & $\times$ & \checkmark \\ \hline\hline

\multirow{3}{*}{TCN} & PyTorch & 0.03451 & 0.03501 & 3.91782  & 1.969 & 1351.8 & $\times$ & $\times$ \\ \cline{2-9}
& LiteRT \texttt{float16} & 0.00877 & 0.00999 & 1.01778 & 0.416 & 363.5 & \checkmark &  $\times$ \\ \cline{2-9}
& LiteRT \texttt{int8} & 0.01113 & 0.01207 & 1.27697 & 0.451 & 362.2 &  $\times$ & \checkmark \\ \hline\hline


\multirow{3}{*}{LSTM} & PyTorch & 0.00832 & 0.00990 & 0.96733 & 2.047 & 831.4 & \checkmark & $\times$ \\ \cline{2-9}
& LiteRT \texttt{float16} & 0.09316 & 0.09586 & 10.60240 & 1.245 & 638.6 &  $\times$ & $\times$ \\ \cline{2-9}
& LiteRT \texttt{int8} & 0.09570 & 0.09847 & 10.89683 & 1.272 & 960.2 & $\times$ & $\times$ \\\hline\hline

\multirow{3}{*}{GRU} & PyTorch & 0.00460 & 0.00513 & 0.53256 & 3.968 & 816.6 & \checkmark & $\times$ \\ \cline{2-9}
& LiteRT \texttt{float16} & 0.00371 & 0.00425 & 0.42814 & 1.080 & 855.9 & \checkmark & $\times$ \\ \cline{2-9}
& LiteRT \texttt{int8} & 0.01807 & 0.02145 & 2.02840 & 1.137 & 1143.6 & $\times$ & $\times$ \\\hline\hline

\multirow{3}{*}{Transformer} & PyTorch & 0.00719 & 0.00763 & 0.83094 & 2.389 & 1327.8 & \checkmark & $\times$ \\ \cline{2-9}
& LiteRT \texttt{float16} & 0.00659 & 0.00719 & 0.76240 & 0.658 & 413.5 &  \checkmark & $\times$ \\ \cline{2-9}
& LiteRT \texttt{int8} & 0.00900 & 0.00969 & 1.03755 & 1.215 & 377.6 & \checkmark & \checkmark \\\hline\hline

\multirow{3}{*}{\shortstack{LNN\\(ours)}} & PyTorch & 0.00781 & 0.00850 & 0.90772 & 1.912 & 616.1 & \checkmark & $\times$  \\ \cline{2-9}
& LiteRT \texttt{float16} & 0.00892 & 0.00950 & 1.03398 & 0.081 & 199.6 & \checkmark & $\times$ \\ \cline{2-9}
& LiteRT \texttt{int8} & 0.00934 & 0.01000 & 1.08152 & 0.132 & 197.4 & \checkmark & \checkmark \\
\hline\hline
\end{tabular}%
}
\label{tab:teacher}
\end{table*}

We first discuss the results for the error objective. We follow a general criterion that a model is well-performed with low errors when its MAE is no larger than 0.01 and we mark these models with $\checkmark$ in the table. We have the following observations. First, LNN is a strong candidate as the teacher model. It begins with a low error level with 0.0078 MAE and its error shift after quantization is small, e.g., 0.0089 and 0.0093 for the \texttt{float16} and \texttt{int8} quantized models, respectively. All these MAEs for LNN are within 0.01, showing it can be a reliable knowledge source. By contrast, alternative ML models exhibit larger errors generally. Among them, CNN and TCN have large errors, e.g., >0.01 MAE, for their non-compressed versions, and are not suitable to be the knowledge source. LTSM and GRU suffer from significantly increased errors after \texttt{int8} quantization. Relatively, Transformer and LNN show competitive accuracy and achieve MAEs below 0.01 for tested implementations.

For the cost objective, we focus on the LiteRT implementations and consider models small enough, i.e., within 500 kB, for edge deployment. LSTM and GRU are not eligible, with GRU's \texttt{int8} version exceeding 1 MB. Besides, when practical settings are considered for edge devices such as Arduino Nano 33 BLE Sense and ESP32, there is no native hardware support for \texttt{float16}. As such, the selection of suitable ML algorithm is highly constrained, as the quantized models cannot be always generated for both \texttt{float16} and \texttt{int8} versions with commonly used model export toolkit. Some models also trigger execution overloads or crashes on compact devices despite performing well on workstations. With both error and cost objectives considered, only Transformer and LNN excel under edge constraints. LNN's MAE is 3.8\% higher than Transformer for \texttt{int8}, but Transformer's size is 91.3\% larger. Overall, the experiment results justify the usage of LNN in \textsc{DLNet} empirically and demonstrate that LNN is a suitable teacher model with good potential for the following distillation and deployment stages. 

\begin{table*}[t]
\centering
\renewcommand{\arraystretch}{1.1}
\setlength{\tabcolsep}{3pt}
\caption{The performance of the student models in the first stage distillation with different hidden state dimensions in $\{2^1,\ldots,2^7\} = \{2,\ldots,128\}$ and \texttt{MSE} and \texttt{Cosine} losses. Error indicators include MAE ($\times 10^{-2}$), RMSE ($\times 10^{-2}$), MAPE (\%), uncertainty ($\times 10^{-3}$), and confidence. Cost indicators include inference time (ms), model size (kB), energy ($\times 10^{-8}$ kWh), and CO2 emission ($\times 10^{-8}$ kg).}
\resizebox{0.98\textwidth}{!}{%
\begin{tabular}{@{}c|c|ccccccc|ccccccc@{}}
\hline\hline
\multirow{2}{*}{\shortstack{Hidden\\Dimension}} & Teacher & \multicolumn{7}{c|}{Students with \texttt{MSE} loss} 
& \multicolumn{7}{c}{Students with \texttt{Cosine} loss} \\
\cline{2-16}
& 128& 128& 64& 32& 16& 8& 4& 2& 128& 64& 32& 16& 8& 4& 2
\\ \hline\hline

MAE & 0.78 & 1.02 & 1.01 & 1.02 & 1.03 & 0.90 & 0.84 & 1.96 & 1.18 & 0.78 & 0.90 & 0.75 & 0.88 & 1.10 & 1.93
\\ \hline
RMSE & 0.85 & 1.08 & 1.07 & 1.08 & 1.10 & 0.97 & 0.90 & 2.09 & 1.22 & 0.84 & 0.94 & 0.81 & 0.95 & 1.14 & 2.03 \\ \hline
MAPE & 0.91 & 1.18 & 1.17 & 1.18 & 1.19 & 1.05 & 0.97 & 2.25 & 1.37 & 0.91 & 1.04 & 0.87 & 1.02 & 1.26 & 2.21 \\ \hline

Uncertainty & 2.50 & 2.69 & 2.78 & 3.03 & 3.22 & 3.20 & 2.87 & 3.53 & 3.35 & 3.07 & 3.28 & 3.76 & 3.78 & 3.27 & 7.73 \\ \hline
Confidence & 58.3 & 50.3 & 56.9 & 58.0 & 56.5 & 61.4 & 58.6 & 26.6 & 48.9 & 61.6 & 60.8 & 64.1 & 63.6 & 47.9 & 46.5
\\ \hline\hline
Inference & 1.91 & 0.76 & 0.98 & 1.09 & 1.03 & 0.94 & 0.93 & 0.82 & 1.02 & 0.90 & 1.04 & 0.97 & 1.06 & 0.94 & 0.93 \\ \hline
Model Size & 616 & 685 & 460 & 371 & 333 & 315 & 306 & 302 & 685 & 460 & 371 & 333 & 315 & 306 & 302 \\ \hline
Energy & 3.90 & 2.69 & 3.06 & 3.21 & 3.20 & 3.12 & 2.97 & 2.86 & 3.10 & 2.83 & 3.17 & 3.04 & 3.20 & 2.93 & 2.95 \\ \hline
CO2 & 1.83 & 1.27 & 1.44 & 1.51 & 1.51 & 1.39 & 1.40 & 1.35 & 1.46 & 1.15 & 1.49 & 1.43 & 1.51 & 1.38 & 1.39 \\ \hline\hline
\end{tabular}%
}
\label{tab:distil-1}
\end{table*}

\subsection{First-Stage Distillation Results}
We investigate whether Euler-based students can inherit the teacher's knowledge while becoming edge-friendly. Here, we generate a student pool with varying hidden state dimension from $2$ to $128$, and we train each student with either \texttt{MSE} and \texttt{Cosine} loss. In total, we produce 14 different students after this stage of distillation. Our goal is not to pick the most accurate model, but to select students with an optimal trade-off between error and cost. We present the experiment results in Table \ref{tab:distil-1}, where the results of different error and cost indicators are available and enable the Pareto-based bi-objective model selection. 

We have the following main observations. First, smaller can win. A student can match or exceed the teacher's accuracy while being smaller and more compact. For example, with \texttt{Cosine} loss and hidden dimension 16, the student improves the teacher model's MAE from 0.0078 to 0.0075. Similarly, the RMSE improves from 0.0085 to 0.0081 and the student reduces the MAPE from 0.91\% to 0.87\%. With such error reductions, the student is also smaller, with around only half of the teacher's model size. Despite such encouraging results, student models that are too small do not perform well. For example, with hidden state dimension 2, errors jump, e.g., from teacher's 0.0078 to the student's 0.0196 and 0.0193 with \texttt{MSE} and \texttt{Cosine} losses, that are 2.51x and 2.47x higher, respectively. The student also has a low confidence, which is only 45.6\% of the teacher's confidence. Besides, the Pareto-optimal students are not always from the same hidden dimension or distillation loss, and this observation aligns with our design preference of introducing diversity when creating the student pool. For example, one optimal student has a hidden dimension of 16 with \texttt{Cosine} loss, and another optimal student is derived with the \texttt{MSE} loss and a different dimension of 4. 

\begin{figure}[t]
   \subfloat[First-Stage Distillation]{\label{fig:pareto1}
      \includegraphics[width=.48\textwidth]{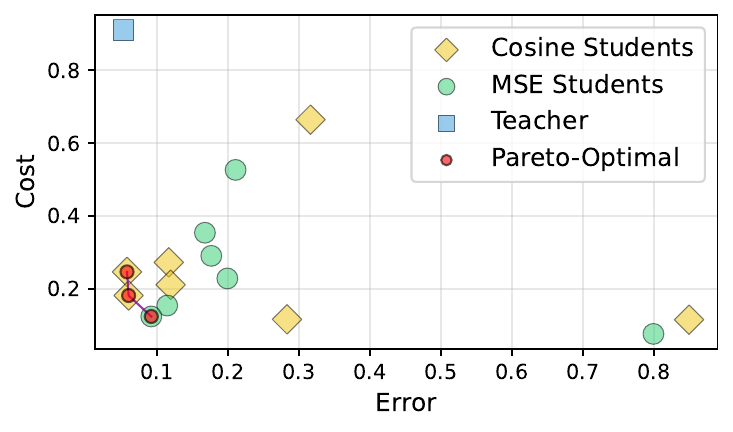}}
~
   \subfloat[Second-Stage Distillation]{\label{fig:pareto2}
      \includegraphics[width=.48\textwidth]{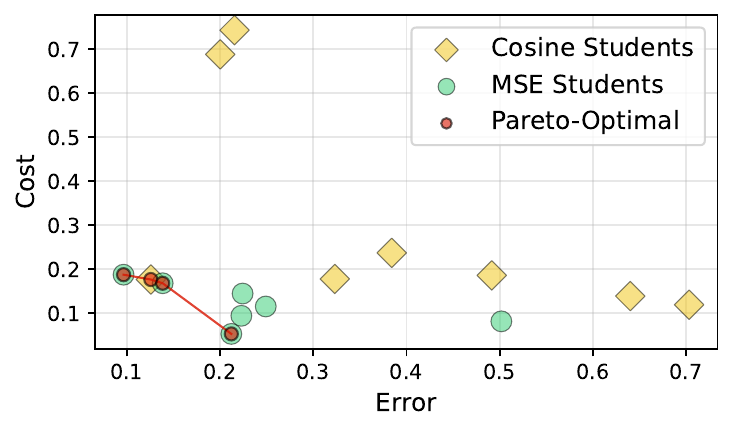}}

   \caption{Illustration of student models from the two distillation stages in the error–cost space. Pareto-selected students with low error-cost are highlighted.}\label{fig:pareto}
\end{figure}

For practical deployment, most students can reduce model size significantly. The same student with dimension 16 and \texttt{Cosine} loss takes only 333 kB, which is 52.0\% of the teacher model size. Reducing the hidden dimension helps obtain more compact models, e.g., from 685 to 302 with 55.9\% size reduction for the students with \texttt{MSE} loss; however, models too small may have high errors as presented above. Also, the results justify that Pareto-based selection is necessary. Different students have different error-cost values, where some students are small but inaccurate, some are accurate but large, and some improve both objectives and are selected as the Pareto-optimal students. Finally, the updated pool includes three students include \texttt{M-4}, \texttt{C-16}, and \texttt{C-64} as shown in Fig. \ref{fig:pareto1}, where the letter and integer refer to \texttt{MSE}/\texttt{Cosine} loss and dimension, respectively. 

\subsection{Second-Stage Distillation Results}
The three students selected from the first distillation undergo pruning with nine sparsity levels from 0.1 to 0.9. and we re-stabilize their performance with another round of distillation. In general, we expect to create smaller student models through pruning while the prediction accuracy remains competitive, and we aim to identify optimal sparsity levels. Among the tested pruned students, some exhibit poor stability, e.g., reporting compilation errors or collapse, especially for the pruned students based on \texttt{C-64}. We report the experiment results of all stable students in Table \ref{table:distil-2}, and we have the following main observations.

\begin{table}[t]
\centering
\renewcommand{\arraystretch}{1.1}
\setlength{\tabcolsep}{3pt}
\caption{The performance of the student models in the second distillation with different sparsity in $\{0.1,\ldots,0.9\}$ based on the three optimal students from the first distillation. Error indicators include MAE ($\times 10^{-2}$), RMSE ($\times 10^{-2}$), MAPE (\%), uncertainty ($\times 10^{-3}$), and confidence. Cost indicators include inference time (ms), model size (kB), energy ($\times 10^{-8}$ kWh), and CO2 emission ($\times 10^{-8}$ kg).}
\resizebox{0.98\textwidth}{!}{%
\begin{tabular}{@{}c|c|ccccccc|cccccc|cc@{}ll}
\hline\hline
Model & Teacher & \multicolumn{7}{c|}{\texttt{M-4}}& \multicolumn{6}{c|}{\texttt{C-16}} & \multicolumn{2}{c}{\texttt{C-64}} \\ \hline
Sparsity & -- & 0.1 & 0.2 & 0.3 & 0.4 & 0.5 & 0.8 & 0.9 
 & 0.1 & 0.2 & 0.3 & 0.4 & 0.5 & 0.9  & 0.1 & 0.2\\ \hline\hline
MAE & 0.78 & 0.76 & 0.91 & 0.99 & 1.08 & 1.02 & 0.86 & 1.30 & 1.31 & 1.48 & 0.81 & 1.23 & 1.78 & 2.20 & 1.01 & 1.03 \\\hline
RMSE & 0.85 & 0.83 & 0.96 & 1.09 & 1.14 & 1.10 & 0.97 & 1.40 & 1.34 & 1.52 & 0.87 & 1.27 & 1.84 & 2.37 & 1.05 & 1.08 \\\hline
MAPE & 0.91 & 0.87 & 1.05 & 1.15 & 1.25 & 1.18 & 0.99 & 1.50 & 1.50 & 1.71 & 0.94 & 1.42 & 2.07 & 2.55 & 1.16 & 1.19 \\ \hline
Uncertainty & 2.50 & 2.98 & 3.13 & 4.07 & 3.42 & 3.28 & 4.89 & 5.81 & 3.67 & 4.29 & 3.59 & 3.14 & 4.74 & 3.24 & 2.57 & 3.58 \\ \hline
Confidence & 58.3 & 58.8 & 61.9 & 60.8 & 60.8 & 58.1 & 68.3 & 57.9 & 49.6 & 49.3 & 64.6 & 49.1 & 47.7 & 43.7 & 49.9 & 61.4 \\ \hline
Inference & 1.91 & 0.70 & 0.78 & 0.73 & 0.87 & 0.61 & 0.92 & 0.87 & 0.87 & 0.82 & 0.94 & 1.00 & 0.91 & 1.02 & 2.21 &2.65 \\ \hline
Model Size & 616 & 273 & 245 & 220 & 198 & 180 & 138 & 128 & 281 & 251 & 223 & 199 & 182 & 143 & 404 &366 \\ \hline
Energy & 3.90 & 2.51 & 2.61 & 2.59 & 2.87 & 2.46 & 2.92 & 2.95 & 3.01 & 2.79 & 2.95 & 3.16 & 2.99 & 3.15  & 6.34 &7.24 \\ \hline
CO2 & 1.83 & 1.18 & 1.16 & 0.75 & 1.29 & 0.64 & 1.37 & 1.39 & 1.34 & 1.19 & 1.24 & 1.49 & 1.34 & 1.42 & 2.98 & 3.41 \\\hline\hline
\end{tabular}%
}
\label{table:distil-2}
\end{table}

First, there are well-performed sparsity levels. For the \texttt{M-4} student, sparsity 0.1 leads to very low errors, e.g., 0.0076 MAE and 0.0083 RMSE, where the model size drops by 10.8\% compared to the original student model. Similar low sparsity produces other competitive students also, e.g., the \texttt{M-4} student with sparsity 0.2 and the \texttt{C-16} student with sparsity 0.3. However, aggressive pruning eventually damages accuracy. For the same \texttt{M-4} student, increasing the sparsity to 0.5, MAE increases from 0.0076 to 0.0102 and RMSE rises to 0.0110. High sparsity helps create small models, which however are not optimal as the error often increases more significantly. Besides, models too small also face the challenge of high uncertainty. For \texttt{M-4}, uncertainty grows from 3.0 at sparsity 0.1 to 5.8 at sparsity 0.9. Uncertainty matters because it indicates whether a model’s prediction is reliable, and real-world deployments, especially for safety-critical systems like BMS, require predictions that are not only accurate but also dependable. We also note that the error-cost indicators often show non-monotonic trends, e.g., increasing sparsity does not consistently worsen MAE. There are fluctuations due to practical reasons such as optimization noises. This is often overlooked in academic studies that assume a definite correlation between pruning and accuracy, and the fact of non-monotonicity highlights the importance of incorporating indicators such as uncertainty into our error objective function. 

Overall, the second stage distillation contributes to the overall model optimization with significant cost reduction and sometimes error improvement also. Finally, as shown in Fig. \ref{fig:pareto2}, there are four optimal students, including the \texttt{M-4} students with sparsity 0.1, 0.2, and 0.5, and \texttt{C-16} with sparsity 0.3. Except \texttt{M-4} with 0.5 sparsity with relatively high errors, the rest share similar performance.

\subsection{Edge Deployment on Arduino Nano}
\begin{figure}[t]
\centering
\includegraphics[width=0.95\textwidth]{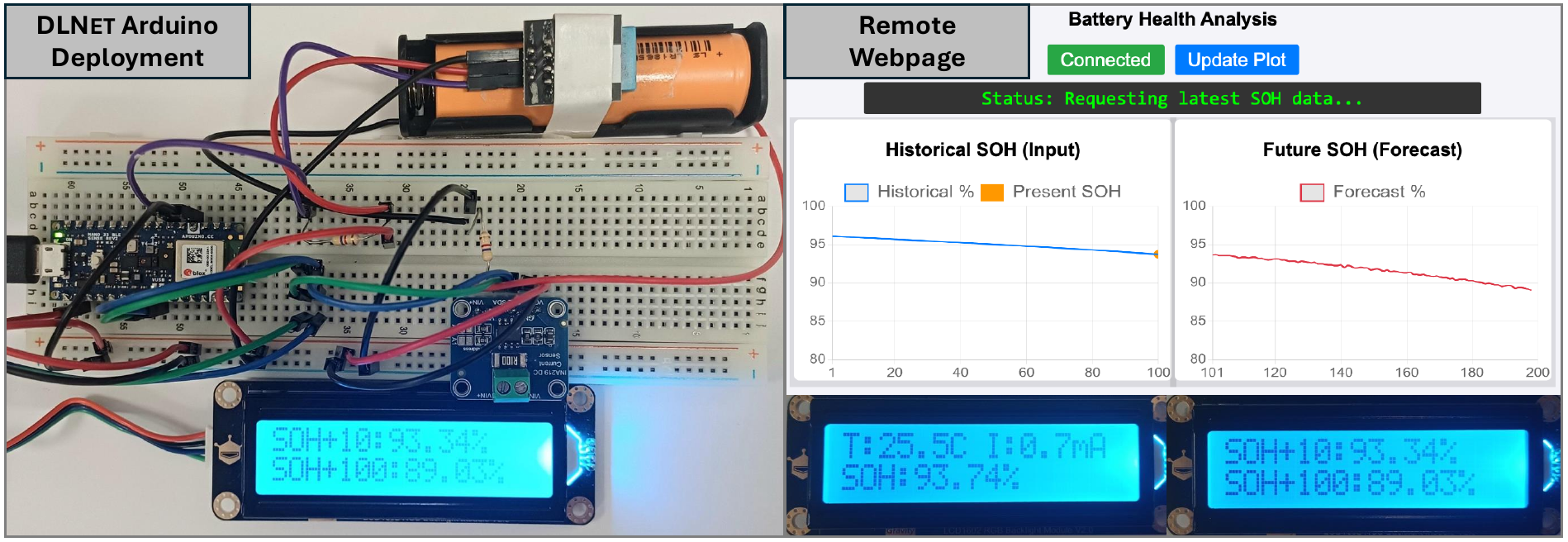}
\caption{A demonstration of \textsc{DLNet} with Arduino-based on-device deployment.}
\label{fig:demo}
\end{figure}

In this part, we demonstrate the effectiveness of \textsc{DLNet}'s edge deployment. We choose one of the optimal students, \texttt{M-4} with sparsity 0.1 for demonstration. We apply post-training \texttt{int8} quantization for the selected student model, export into a LiteRT-compatible graph, and convert the exported model into an embedded C representation. Arduino is chosen as the edge platform because it is widely used in embedded prototyping and represents more compact deployments than typically single-board edge devices such as Raspberry Pi. However, not all Arduino boards support on-device ML inference, and we select Arduino Nano 33 BLE Sense as a relatively compact option with ML support. Once feasibility is demonstrated on such a constrained board, edge deployments can be extended to other compact devices. The board is a Nordic \texttt{nRF52840} BLE microcontroller with 1 MB on-chip flash memory and 4 kB page size. We allocate a fixed tensor arena, load the model, register the required operators, and run inference with the same input window and forecast horizon configurations as the teacher model. Finally, we validate the deployment and report results in Table \ref{tab:edge} and Fig. \ref{fig:demo}.

\begin{table}[t]
\centering
\scriptsize
\caption{Comparison between the teacher on a workstation and \textsc{DLNet} on an Arduino Nano MCU. Error metrics including MAE, RMSE, MAPE, MAE$_{10}$, and MAE$_{100}$ are reported in $\times 10^{-2}$. Efficiency metrics include memory (MB), model size (kB), FLOPs (M), parameter count ($\times 10^{3}$), and inference time (ms). The gap row reports relative change (\%) of \textsc{DLNet} with respect to the teacher.}
\label{tab:edge}
\setlength{\tabcolsep}{2pt}
\renewcommand{\arraystretch}{1.1}
\resizebox{0.98\columnwidth}{!}{%
\begin{tabular}{@{}c|ccccc|ccccc@{}} \hline \hline
Model & MAE & RMSE & MAPE & MAE$_{10}$ & MAE$_{100}$ & Memory & Model Size & FLOPs & \# Paras & Inference \\ \hline
Teacher & 0.78 & 0.85 & 0.91 & 0.51 & 1.28 & 640.43 & 616.1 & 0.288 & 155.8 & 1.9 \\ \hline
\textsc{DLNet} & 0.66 & 0.73 & 0.76 & 0.43 & 1.03 & 0.02 & 94.1 & 0.068 & 70.5 & 21.1 \\ \hline
Gap & ↓ 15.4 & ↓ 13.9 & ↓ 15.8 & ↓ 15.6 & ↓ 19.4 & ↓ 100.0 & ↓ 84.7 & ↓ 76.4 & ↓ 54.7 &
↑ 1002 \\ \hline \hline
\end{tabular}%
}
\end{table}

The final \textsc{DLNet} student deployed on Arduino improves prediction performance over the teacher on a workstation. The MAE over the full forecast horizon drops from 0.0078 to 0.0066, corresponding to a 15.4\% error reduction compared to the teacher. RMSE and MAPE also decrease with comparable error reductions. We further report errors of cycle-10 and cycle-100 SoH predictions. As expected, \textsc{DLNet} is more accurate in the near term, with 0.0043 MAE at cycle-10, and the error increases to 0.0103 at cycle-100. This suggests that, under strict accuracy requirements, practical systems benefit from regular forecasting updates rather than relying on a single long-horizon prediction. Notably, the improvement margin is larger at cycle-100 than cycle-10, indicating that \textsc{DLNet} reduces accumulated drift and enhances long-horizon reliability. On the deployment-cost, \textsc{DLNet} achieves a substantial reduction in memory footprint from 640 to 0.02 MB, which enables microcontroller-level feasibility. The model size also shrinks from 616 to below 100 kB. The teacher achieves low inference latency on the workstation, which is expected given much more compute resources. Nevertheless, \textsc{DLNet} completes one inference in 21 ms on Arduino, which is acceptable for battery analytics since degradation evolves slowly over long time scales. When evaluated under the same workstation, \textsc{DLNet} runs in 0.14 ms, 7\% of the teacher’s time only. Overall, we provide evidence that smaller can win in practice, where \textsc{DLNet} improves accuracy while substantially reducing cost.

\section{Conclusion}
\label{sec:conclusion}
In this paper, we presented \textsc{DLNet}, a deployment-aware framework that turns a high-capacity teacher into compact and edge-ready students for battery health prognostics. First, \textsc{DLNet} builds a diverse set of students, distills the teacher's knowledge into them, and keeps the Pareto-optimal candidates that balance accuracy and efficiency. Then, it prunes these optimal students and performs another round of distillation to restore useful teacher behavior after pruning, and re-selects the final practical students for deployment. We validated \textsc{DLNet} on the widely used MIT-Stanford battery dataset and demonstrated real-device feasibility on an Arduino Nano 33 BLE Sense with \texttt{int8} quantization. The final deployed student achieves 0.0066 MAE for predicting the SoH across the future 100 cycles, corresponding to 15.4\% error reduction compared to the teacher model. \textsc{DLNet} enables a compact model, reducing the teacher's 616 to 94 kB with 84.7\% size reduction. These results highlight that, with proper supervision and selection, compact models can be both accurate and deployable for real-world battery management. In addition, \textsc{DLNet}'s framework is designed with flexibility and can extend to other resource-constrained industrial applications.

\subsubsection{Acknowledgments} 
This work was supported in part by A*STAR under its MTC Programmatic (Award M23L9b0052), MTC Individual Research Grants (IRG) (Award M23M6c0113), and MTC Programmatic (Award M24N6b0043), and the Shanghai Sci-tech Co-research Program (Award 25HB2702600).

\bibliographystyle{splncs04}
\bibliography{references}

\end{document}